\documentclass[letterpaper, 10 pt, conference]{ieeeconf}  
\usepackage{graphicx} 
\usepackage{amsmath} 
\usepackage{flushend} 

\IEEEoverridecommandlockouts                              
\title{\LARGE \bf
Extracting Cellular Location of Human Proteins Using Deep Learning
}

\author{Hanke Chen\\
Sandy Spring Friends School\\
16923 Norwood Rd, Sandy Spring, MD 20860\\
{\tt\small hanke.chen@ssfs.org; i@chenhanke.me}
}

\begin{document}

\maketitle
\thispagestyle{empty}
\pagestyle{empty}

\begin{abstract}
    
Understanding and extracting the patterns of microscopy images has been a major challenge in the biomedical field. Although trained scientists can locate the proteins of interest within a human cell, this procedure is not efficient and accurate enough to process a large amount of data and it often leads to bias. To resolve this problem, we attempted to create an automatic image classifier using Machine Learning to locate human proteins with higher speed and accuracy than human beings.
We implemented a Convolution Neural Network with Residue and Squeeze-Excitation layers classifier to locate given proteins of any type in a subcellular structure. After training the model using a series of techniques, it can locate thousands of proteins in 27 different human cell types into 28 subcellular locations, way significant than historical approaches. The model can classify 4,500 images per minute with an accuracy of 63.07\%, surpassing human performance in accuracy (by 35\%) and speed.
Because our system can be implemented on different cell types, it opens a new vision of understanding in the biomedical field. From the locational information of the human proteins, doctors can easily detect cell's abnormal behaviors including viral infection, pathogen invasion, and malignant tumor development. Given the amount of data generalized by experiments are greater than that human can analyze, the model cut down the human resources and time needed to analyze data. Moreover, this locational information can be used in different scenarios like subcellular engineering, medical care, and etiology inspection.

\end{abstract}

\section{Introduction}

So far, the research in protein classification is limited to finding a distinct pattern in a single or a few cell types. Besides, the current method of localizing proteins by hand is time-consuming and may lead to subjective bias. These limitations prohibit the further understanding of the protein distribution within different types of cell. Since the current research on human protein is inefficient and there are a large amount of data remain unanalyzed, a method to speed up the research progress is required.
Our objective is to train and improve a Convolution Neuron Network using subcellular images and compare its result in speed and accuracy with of our performance after training. We hypothesize that the modern Machine Learning approach can correctly classify the proteins into different subcellular locations with accuracy greater than an ordinary trained human performance and speed of less than 1 sec/image. The subcellular images are generally hard to interpret even for trained citizen scientists. Letting a machine to do the work is even more challenging. During the experiment, we would spend most of the time iterating the generations of the model to get the optimal accuracy. The result is expected to surpass human behavior just by 10\%.

\section{Related Work}
\subsection{An Approach for HeLa Single Cell}
In 2007, Chebria's team used Machine Learning to recognize and classify the major subcellular locations using 2D HeLa single-cell images dataset. They extracted multiresolution (MR) features and trained a basic two-layer Neuron Network on the dataset. Their accuracy on the dataset with a set of multiresolution features was 95.3\% (Chebira et al., 2007).
However, the model they trained on the 2D HeLa single-cell images dataset does not generalize to other cell types with more than one cell in an image.

\subsection{An Approach for Human Reproductive Tissue}
In 2012, Fan Yang, Ying-Ying Xu, and Hong-Bin Shen published another gradient based classifier to classify 7 major subcellular classes using the Human Protein Atlas (HPA) database. They selected important feature subsets such as wavelet Haralick features and local binary patterns and ensembled the models. After all, the model achieved 84\% accuracy on the test set and 98\% accuracy on the most confident classifications (Yang, Xu, \& Shen, 2012).
This research successfully classified and located different human proteins in reproductive tissues. But more researches are required for obtaining meaningful information from the location of the protein in other tissues.

\subsection{An Approach for Yeast Proteins}
In 2017, experimenters in the University of Tartu also used an 11-layer neural network to classify the subcellular locations of yeast proteins. They classified two channel images (denoting protein of interest and subcellular location) into 12 subcellular zones using basic Convolution Neuron Network (CNN) architecture as the feature extractor. The model achieved 91\% per cell classification accuracy, and 99\% per protein accuracy (T \& L, 2017).
Although this approach established a baseline locating the yeast proteins, the structure of yeast cells is relatively simple compared to human cells of different types, making it unsuitable to generalize the distributional patterns of human proteins.

\subsection{Our Approach}
Inspired by the articles, our objective is to go a step further to generalize the solution to more proteins in more tissues with more subcellular categories without the limitation to specific tissues, cells or proteins. The method we present would localize proteins in 27 different human cell types with 28 different subcellular locations, far greater than the historical approaches above.
Subcellular protein distribution may reflect the current status of a cell, making the classification of cells with abnormal cell growth easier. For example, exportin-1 (XPO1), or chromosomal region maintenance 1 (CRM1), transport other proteins between the cytoplasmic area and the nucleus to maintain the normal functions of a cell (Parikh, Cang, Sekhri, \& Liu, 2014). When there is an imbalance of proteins distribution inside the nucleus, the chance of viral replication, inflammatory development, and malignant tumor transformation would also increase (Cast Pharma, 2015). This means that by correctly localize proteins in human cells, researches can make use of a large number of unanalyzed microscopy images to identify various types of infection within a cell.

\section{Training and Inference}
\subsection{Data Analysis and Pre-Processing}

\begin{figure}[h!]
  \centering
  A Sample of HPA Dataset in Training Set\par\medskip
  \includegraphics[width=0.45\textwidth]{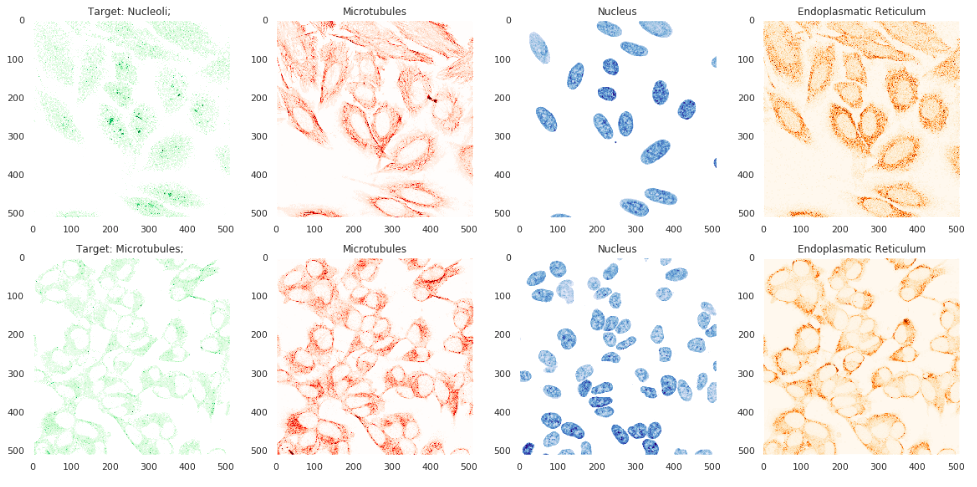}
  \caption{These are two samples of the unprocessed microscopic image from the dataset. The dataset consists of 31072 training images and 11702 testing images with four channel, denoting the cellular location (red, blue, and yellow) and proteins of interest (green).}
\end{figure}

\begin{figure}[h!]
  \centering
  The Distribution Graph of Training and Validation Data Split\par\medskip
  \includegraphics[width=0.45\textwidth]{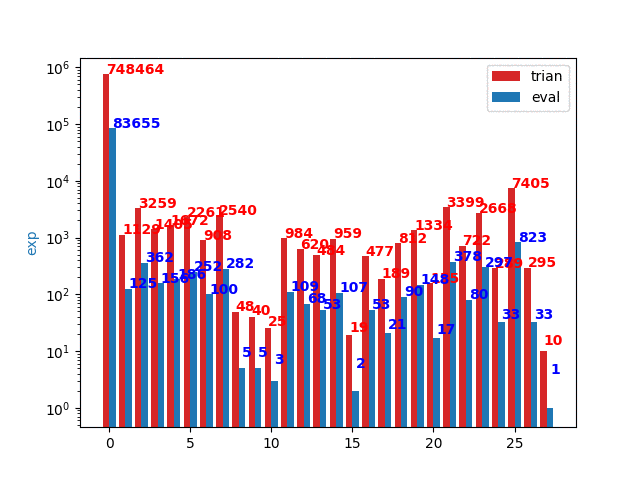}
  \caption{Since the dataset is extremely unbalanced, we used categorical stratification to split the training and testing data. The above image is the split distribution for fold 1.}
\end{figure}


In this experiment, we used the Human Protein Atlas (HPA) Image Classification data on Kaggle and an additional HPA v18.1 dataset for testing. HPA dataset has 1000:1 data unbalance which may result in the gradient explosion problem when using unweighted loss functions. We split the data into 10 folds cross-validation by stratifying using the label (The Human Protein Atlas, n.d.).

\begin{figure}[h!]
  \centering
  Image Augmentation Method\par\medskip
  \includegraphics[width=0.45\textwidth]{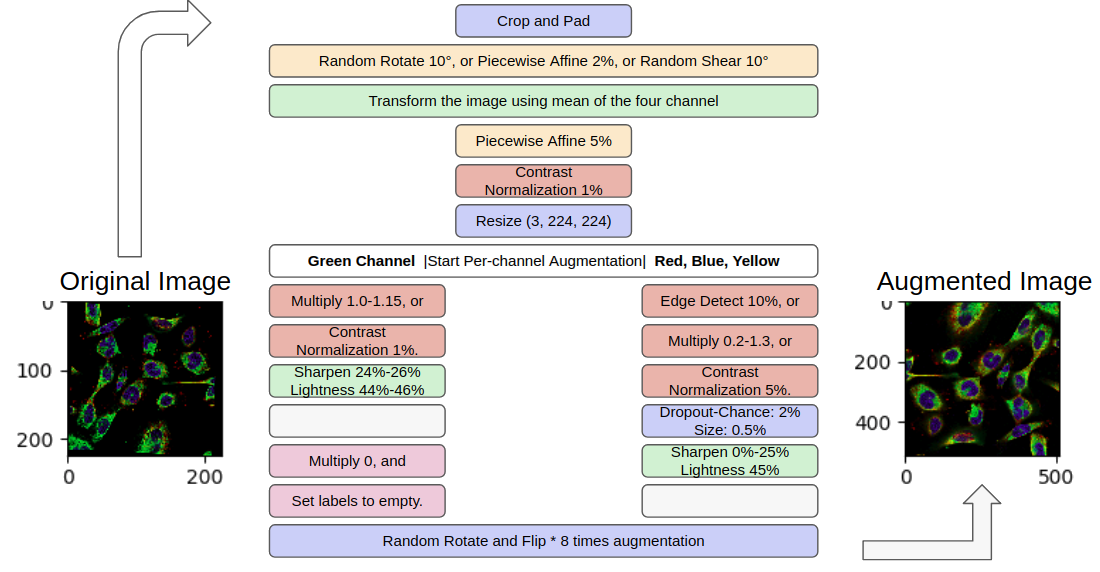}
  \caption{These are the image pre-processing method we use to complete the full distribution of possible images. The blue operations are basic operations that remove the pixel values. The orange operations rotate and shift pixel values. The red operations adjust the pixel values. The green operations also adjust the pixel values, but this is added to fix the blurry effect as the result of resizing image. The purple operations create negative samples from the dataset. Only one of the parallel operations are chosen for each image in one epoch.}
  \label{image-augmentation-method}
\end{figure}

We augmented the training set randomly based on our augmentation algorithm as showed above in \ref{image-augmentation-method}.

\subsection{Network Architecture}

\begin{figure}[h!]
  \centering
  SE-ResNext Block\par\medskip
  \includegraphics[width=0.45\textwidth]{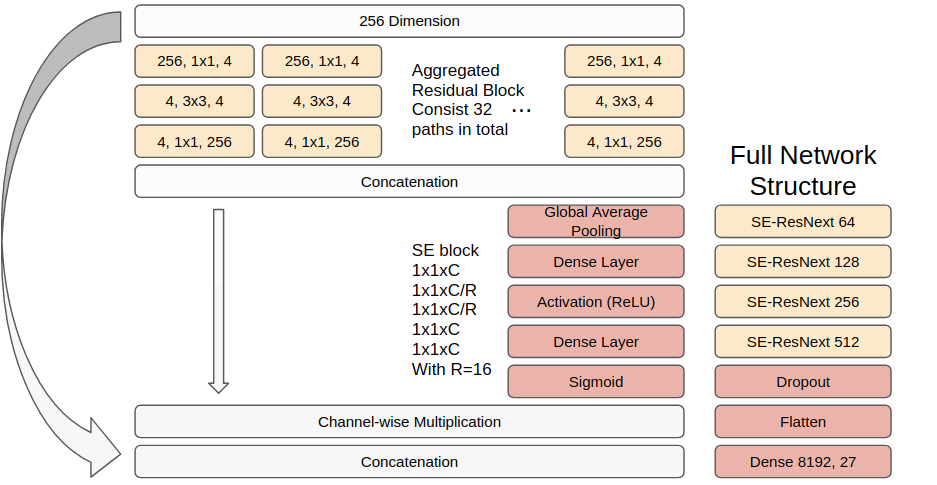}
  \caption{This image demonstrates the SEResNext architecture we use for the experiment. The Aggregated Residual Block allowed the gradient to flow from high to low layers and it was inspired by Inception blocks. The SE Block adds channel-wise inter-dependency and selects more important features by multiplication. It is crucial in protein localization problems because it allows more communications between the green channel and other channels (Kaiming, Zhuowen, Piotr, Ross, \& Saining, 2016).}
\end{figure}

Our network consists of ideas from Aggregated Residual Block from ResNext (Kaiming, Zhuowen, Piotr, Ross, \& Saining, 2016). We also added Squeeze and Excitation Block from SENet (Hu, Shen, Albanie, Sun, \& Wu, 2017).

\subsection{Training Process}

\begin{figure}[h!]
  \centering
  Loss and Score Values During Training Process\par\medskip
  \includegraphics[width=0.45\textwidth]{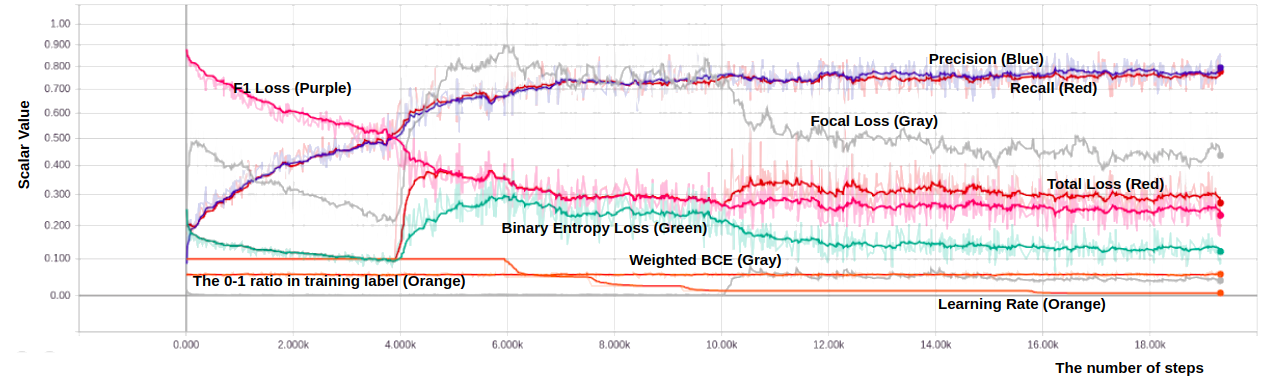}
  \caption{The image above shows 9 different losses and scores we keep tracking during the training process. However, most of them are not used by back-propagation. The sum of the loss we used in back-propagation is the Total Loss. Others are used to track the performance of the model in each step. Notice around step=4k, we switch the loss from BCE to combination of Weighted BCE and F1 Loss. At step=10k, we scaled up the Weighted BCE so it would have a greater impact of the back-propagation process.}
\end{figure}

We trained the network on 16 GPU with 48 GB memory and a Nvidia Tesla P100 CPU for more than 24 hours using PyTorch. To get a faster startup, we used pre-trained SE-ResNext on ImageNet (Russakovsky et al., 2009). During the training process, we recorded several different losses. For the first stage of the training (10 epochs), we used Binary Entropy Loss (BCE) to warm up the gradient. (If we use the focal loss from the beginning, the precision would reach 99\% while the recall would remain low.) After the first 10 epochs, we switch to the combination of soft F1 Loss and negatively weighted BCE. We used a batch size of 64 with an initial learning rate of 0.1 on Adadelta optimizer. Instead of decreasing learning rate each epoch, it is more reasonable to decrease it on each step consider the size of the dataset. Based on the training F1 loss, We decrease the learning rate on the plateau by a factor of 0.5 with initial learning rate set up to 0.1. We used the rest of the evaluation data for epoch evaluation. We implemented 4 times test time augmentation (TTA) so that the result would be more stable and precise. We stopped training after seeing a horizontal fluctuation of the evaluation metrics for 5 epochs. The final training F1 Score is 0.75 with the precision of 77.28\% and the recall of 75.28\%.

\subsection{Post-Processing}

\begin{table}[h!]
\centering
\caption{Best Threshold of Each Subcellular Location (Class)}
\resizebox{0.45\textwidth}{!}{%
\begin{tabular}{|l|l|l|}
\hline
Class & Best Threshold (Raw) & Best Threshold (Smoothed) \\ \hline
All   & 0.2332               & 0.2196                    \\ \hline
0     & 0.07007              & 0.1547                    \\ \hline
1     & 0.965                & 0.1571                    \\ \hline
2     & 0.8579               & 0.1798                    \\ \hline
3     & 0.1662               & 0.1931                    \\ \hline
4     & 0.7728               & 0.1324                    \\ \hline
5     & 0.01001              & 0.1926                    \\ \hline
6     & 0.01201              & 0.09215                   \\ \hline
7     & 0.003003             & 0.1843                    \\ \hline
8     & 0.7978               & 0.1669                    \\ \hline
9     & 0.01602              & 0.09612                   \\ \hline
10    & 0.1982               & 0.01602                   \\ \hline
11    & 0.5325               & 0.1286                    \\ \hline
12    & 0.2152               & 0.1722                    \\ \hline
13    & 0.03103              & 0.1544                    \\ \hline
14    & 0.004004             & 0.04645                   \\ \hline
15    & 0.04304              & 0.06961                   \\ \hline
16    & 0.005005             & 0.1499                    \\ \hline
17    & 0.003003             & 0.06373                   \\ \hline
18    & 0.0981               & 0.1001                    \\ \hline
19    & 0.04204              & 0.1706                    \\ \hline
20    & 0.01101              & 0.1264                    \\ \hline
21    & 0.01101              & 0.1121                    \\ \hline
22    & 0.01702              & 0.08679                   \\ \hline
23    & 0                    & 0                         \\ \hline
24    & 0.03504              & 0.08634                   \\ \hline
25    & 0.01502              & 0.1221                    \\ \hline
26    & 0.005005             & 0.1943                    \\ \hline
27    & 0.01502              & 0.118                     \\ \hline
\end{tabular}%
}\par
\medskip The above image shows the optimal threshold for each class in the last epoch. Notice that the raw thresholds have a lot of noise. The smoothed thresholds are calculated using the best thresholds from the last few epochs. \par
\label{best-threshold-of-each-subcellular-location-class}
\end{table}

Because we discovered the F1 Score of major classes plateaued from the threshold of 0.1 to 0.9 and the best thresholds from the last epoch contains a lot of noise, we hand-picked a maximum value, 0.268, using its score from the evaluation set.

\section{Results}
We used HPA v18 dataset as the test set since it was unused in both the training and the validation process. Because 11,111 images are used for testing the model, the results below have high precision.

The following formula is used for calculating the F Score for validation and testing. Beta here is chosen to be 1.
$$F_\beta=\frac{(1+\beta^2)\cdot \text{TP}}{(1+\beta^2)\cdot \text{TP} + \beta^2 \cdot \text{FN}+\text{FP}}$$

\begin{figure}[h!]
  \centering
  The Validation F1 Score and Focal Score by Number of Epochs Trained\par\medskip
  \includegraphics[width=0.45\textwidth]{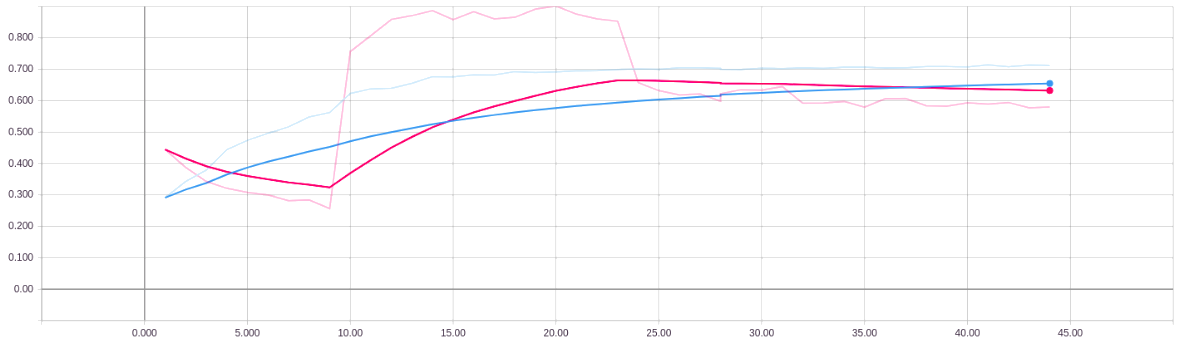}
  \caption{The image above shows the F1 Score (Blue) and Focal Score (Red) of the model in validation. Although Focal loss can reach 0.9 easily, the experiment finds out that it does not representative of the overall performance of the model. The F1 Score shows a general improvement as more epochs are trained.}
\end{figure}

\begin{figure}[h!]
  \centering
  The Validation F1 Score by Number of Epochs Trained\par\medskip
  \includegraphics[width=0.45\textwidth]{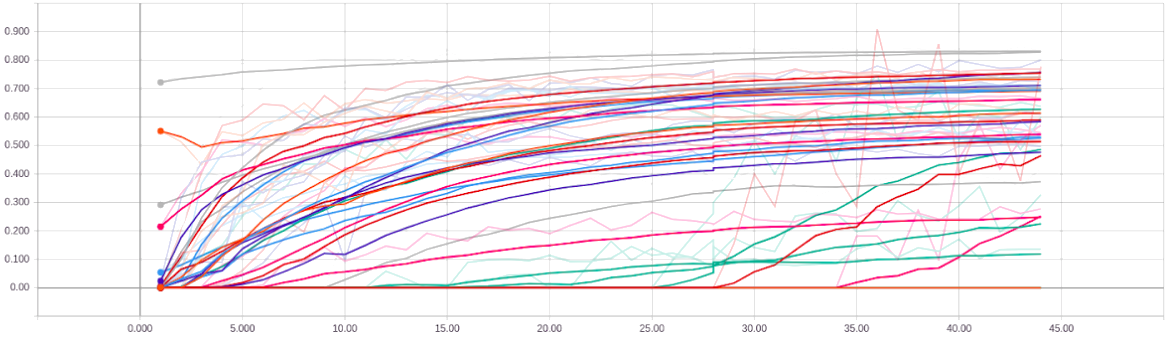}
  \caption{The image above shows the F1 Score of each individual classes. It demonstrates an overall performance for most of the classes. Because of the imbalance in the dataset, some classes have low accuracy compared to others.}
\end{figure}

\begin{figure}[h!]
  \centering
  The Precision vs. Recall Curve of the Final Model\par\medskip
  \includegraphics[width=0.45\textwidth]{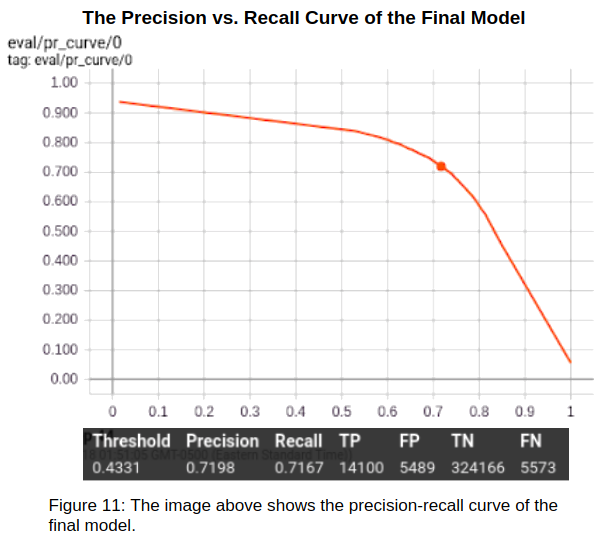}
  \caption{The image above shows the precision-recall curve of the final model.}
\end{figure}

\begin{figure}[h!]
  \centering
  Machine Performance vs. Human Performance\par\medskip
  \includegraphics[width=0.45\textwidth]{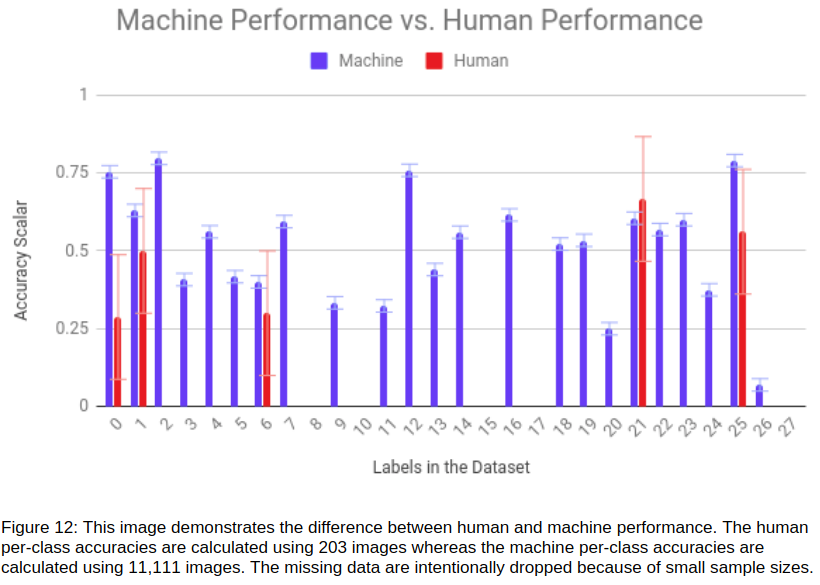}
  \caption{This image demonstrates the difference between human and machine performance. The human per-class accuracy are calculated using 203 images whereas the machine per-class accuracy are calculated using 11,111 images. The missing data are intentionally dropped because of small sample sizes.}
\end{figure}

We simply trained ourselves on the 28 subcellular structures for 2 days and located 203 images for comparison.

\begin{table}[h!]
    \centering
    \caption{Human Accuracy vs. Model’s Accuracy}
    \resizebox{0.45\textwidth}{!}{%
    \begin{tabular}{|l|l|l|}
    \hline
                    & Human   & Machine \\ \hline
    Correct Label   & 5360    & 301384  \\ \hline
    Total Label     & 5880    & 311108  \\ \hline
    Binary Accuracy & 91.15\% & 96.87\% \\ \hline
    F1-Macro        & 0.1124  & 0.3407  \\ \hline
    Precision       & 44.67\% & 67.29\% \\ \hline
    Recall          & 27.46\% & 69.23\% \\ \hline
    IOU             & 27.29\% & 63.07\% \\ \hline
    \end{tabular}%
    }\par
    \medskip This table summarizes the machine’s better performance than humans. The model has about 35\% more IOU accuracy than a human.\par
    \label{human-accuracy-vs-models-accuracy}
\end{table}

\begin{table}[h!]
    \centering
    \caption{Comparison of the Model’s Performance in Speed}
    \resizebox{0.45\textwidth}{!}{%
    \begin{tabular}{|l|l|l|l|}
    \hline
    Image Size       & 4x1728x1728    & 4x512x512      & 4x512x512      \\ \hline
    Batch Size       & 1              & 64             & 1              \\ \hline
    Format           & .jpg           & .npy           & .npy           \\ \hline
    GPU (Nvidia)     & 1x P100        & 1x P100        & 1x P100        \\ \hline
    CPU              & 16 vCPU        & 16 vCPU        & 16 vCPU        \\ \hline
    Speed (/img)     & 1.16s          & 0.0128s        & 0.0769s        \\ \hline
    \end{tabular}%
    }\par
    \medskip The speeds of image processing using our models are calculated using 11,111 sample images. With the same accuracy, the smaller image size achieves higher speed. If the image is pre-processed to .npy file and process with larger batch size, the speed would increase by magnitudes. Using the information above, it is expected that our model can locate around 4,600 images per minute.\par
    \label{comparison-of-the-models-performance-in-speed}
\end{table}

\subsection{Performance Data Summary}

\begin{figure}[h!]
  \centering
  The Number of Generation of Our Experimental Models vs. Training F1 Score\par\medskip
  \includegraphics[width=0.45\textwidth]{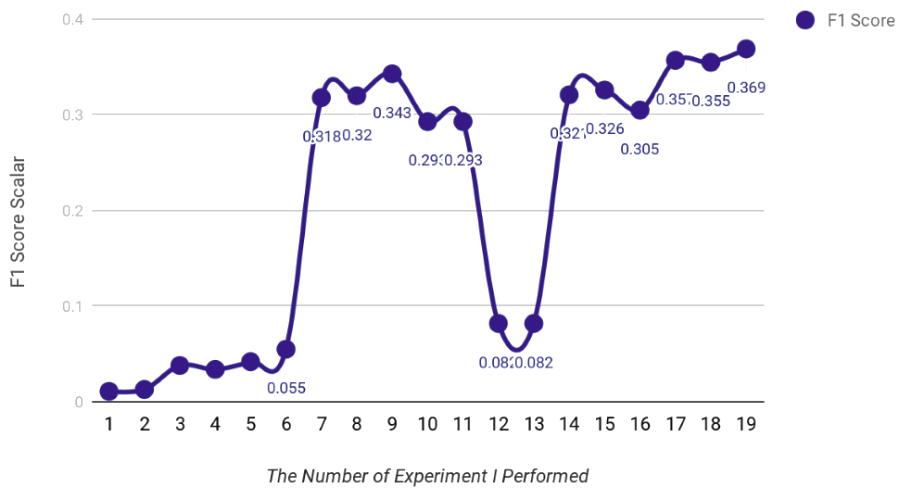}
  \caption{This image demonstrates the overall performance of each generation of our experimental models. This positive trend shows the importance of iterating the product.}
\end{figure}

After 40 epochs trained on 90\% of HPA Kaggle dataset, the F1 score reaches its maximum value. The accuracy of our model is 63.07\% with binary accuracy 96.87\% (compared to human accuracy of 27.29\% with binary accuracy 91.15\%), while the F1 Score of our model is 0.3407. With the most confident class, the model can reach the accuracy of 79.68\%. Within the subcellular locations, the Nucleoli has the highest accuracy of 79.68\% among all classes due to more available data for the class. In general, the machine can reach 35\% higher accuracy than a trained human. Besides the accuracy, the result shows that the model's processing speed (78.125 images per second) is way faster than that of a human.

\section{Conclusion}
In the experiment, we used HPA dataset to train, validate, and test the model we created. The major challenges we encountered are the lack of training power, 1000:1 label imbalance, and 4-channel image processing. Despite the difficulties, the ideas of the Residual block, Squeeze and Excitation block as well as the techniques like augmentation, k-fold cross-validation, test time augmentation, loss selection, hyper-parameter tuning, post-processing, and threshold selection all contributed to the success of the model.
In the end, our model is able to surpass human performance both in speed and accuracy. The final generation of the model reaches 63.07\% accuracy, 96.87\% binary accuracy, surpassing human accuracy by 35\%. Besides, the model is capable of processing around 4,500 images per minute in the experimental condition. These achievements suggest that the experiment is successful and our assumption was correct: our model can correctly classify the proteins into different subcellular locations with accuracy greater than an ordinary trained human performance and speed of less than 1 sec/image.
After training the model, the model can generalize the subcellular locations of any proteins in 28 different human cell types. The model surpasses the human performance both in speed and accuracy. Because of these achievements, scientists can now perform large-scale data analyzation to make use of the overflowing laboratory data, producing useful insights about subcellular structure. From the locational information of the human proteins, doctors can easily detect cell's abnormal behaviors including viral infection, pathogen invasion, and malignant tumor development. The locational information can also be used in scenarios like subcellular engineering, medical care, and etiology inspection.

\section*{Acknowledgement}
This research paper is supported by the Google Science Fair. We thank Google Cloud Platform for providing computational power required in the researching process. We thank teachers in Sandy Spring Friends School who are willing to take a look of the draft. We thank Human Protein Atlas for providing all the image data.

\addtolength{\textheight}{-12cm}   


\end{document}